\documentclass{article} 
\usepackage{iclr2025_conference,times}

\usepackage{amsmath,amsfonts,bm}

\def\eqref#1{equation~\ref{#1}}

\def\1{\bm{1}}

\DeclareMathAlphabet{\mathsfit}{\encodingdefault}{\sfdefault}{m}{sl}
\SetMathAlphabet{\mathsfit}{bold}{\encodingdefault}{\sfdefault}{bx}{n}

\usepackage{newunicodechar}
\newunicodechar{，}{,}

\usepackage{hyperref}
\usepackage{url}
\usepackage{amsmath,amssymb}
\usepackage{booktabs}
\usepackage{multirow}
\usepackage{adjustbox}
\usepackage{amssymb}
\usepackage{wasysym}
\usepackage{microtype}
\usepackage{booktabs}
\usepackage{xcolor}
\usepackage{tabularx}
\usepackage{wrapfig}
\usepackage{CJK}
\usepackage[utf8]{inputenc}

\usepackage{inconsolata}
\usepackage{amsmath}
\usepackage{hyperref}
\usepackage{url}
\usepackage{graphicx}
\usepackage{multirow}
\usepackage{makecell}
\usepackage{booktabs}
\usepackage{amssymb}
\usepackage{color}
\usepackage{adjustbox}
\usepackage{algorithm}
\usepackage{amsmath} 
\usepackage{algorithmic}
\usepackage[most]{tcolorbox}
\usepackage{tablefootnote}
\usepackage{caption}
\usepackage{graphicx}
\usepackage{subcaption}
\usepackage{svg}

\usepackage{enumitem}

\newcommand{\model}{Libra-RM }
\newcommand{\bench}{Libra Bench }

\newcommand{\modelnospace}{Libra-RM}
\newcommand{\benchnospace}{Libra Bench}

\title{\includegraphics[height=1.0em]{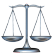} 
Libra: Assessing and Improving Reward Model by Learning to Think}

\iclrfinalcopy

\author{\textbf{Meng Zhou, Bei Li, Jiahao Liu, Xiaowen Shi, Yang Bai, 
Rongxiang Weng\thanks{\llap{}\:\:\: Corresponding author.}}, \\
\textbf{Jingang Wang, Xunliang Cai} \\
Meituan \\
\texttt{\{zhoumeng19,libei17,liujiahao12,shixiaowen03,baiyang28,} \\
\texttt{wengrongxiang,wangjingang02,caixunliang\}@meituan.com}
}

\begin{document}
\maketitle
\begin{abstract}
Reinforcement learning (RL) has significantly improved the reasoning ability of large language models. 
However, current reward models underperform in challenging reasoning scenarios and predominant RL training paradigms rely on rule-based or reference-based rewards, which impose two critical limitations:
1) the dependence on finely annotated reference answer to attain rewards;
and 2) the requirement for constrained output format.
These limitations fundamentally hinder further RL data scaling and sustained enhancement of model reasoning performance.
To address these limitations, we propose a comprehensive framework for evaluating and improving the performance of reward models in complex reasoning scenarios.
We first present a reasoning-oriented benchmark (\benchnospace), systematically constructed from a diverse collection of challenging mathematical problems and advanced reasoning models, to address the limitations of existing reward model benchmarks in reasoning scenarios. 
We further introduce a novel approach for improving the generative reward model via learning-to-think methodologies. Based on the proposed approach, we develop \model series, a collection of generative reward models with reasoning capabilities that achieve state-of-the-art results on various benchmarks.
Comprehensive downstream experiments are conducted and the experimental results demonstrate the correlation between our \bench and downstream application, and the potential of \model to further improve reasoning models with unlabeled data.\footnote{Our \bench is available at \url{https://huggingface.co/datasets/meituan/Libra-Bench}.}
\end{abstract}

\section{Introduction}
\label{sec:intro}
Recent advances in reinforcement learning (RL) and inference-time scaling have significantly unlocked the potential of large language models (LLMs), greatly enhancing their reasoning capabilities~\citep{openai2024learning, guo2025deepseek}. Unlike reinforcement learning from human feedback (RLHF)~\citep{ouyang2022training, bai2022training, zheng2023secrets, xiong2023iterative}, the current RL training paradigms for reasoning models predominantly rely on rule-based or reference-based reward~\citep{guo2025deepseek, yang2025qwen3,seed2025seed, lambert2024t}.
Despite high accuracy, these methods rely on a finely annotated reference answer to attain rewards and a constrained output format to extract the key answer, which limit the use of large-scale data for general reinforcement learning.

To overcome these limitations, there is an urgent need to re-evaluate and advance the role of Reward Models (RMs) as robust proxies for human judgment, especially for general, unlabeled, or hard-to-standardize data. However, existing RMs and their associated benchmarks fall short in complex reasoning scenarios due to three key aspects:
1) existing RM benchmarks are insufficient to assess reward models in complex reasoning scenarios, due to the absence of challenging questions and responses from advanced reasoning models ~\citep{lambert2024rewardbench,frick2024evaluate,zhou2024rmb,tan2024judgebench,zheng2024processbench, song2025prmbench, liu2024rm}; 2) current RMs are designed without deep thinking capabilities and exhibit limited effectiveness when dealing with complex problems ~\citep{tan2024judgebench, zheng2024processbench, acemath2024}; 3) traditional pairwise comparison learning objective of RMs does not align with the correctness metrics in reasoning tasks ~\citep{yang2024qwen25mathtechnicalreportmathematical, acemath2024}.

To address these limitations, we propose a comprehensive framework for evaluating and improving the performance of reward models in challenging reasoning scenarios. We first present a reasoning-oriented RM benchmark, named \bench, to alleviate the shortcomings of existing RM benchmarks. 
The \bench is curated from a diverse collection of challenging mathematical problems and advanced reasoning models, and aims to assess pointwise judging accuracy in terms of correctness. 
These characteristics collectively ensure that our \bench is well aligned with current research and development of reasoning models. 
Through our \benchnospace, we clearly observe and analyze the limitations of existing RMs in challenging reasoning scenarios.

Based on these observations, we further introduce a novel approach for improving the generative reward model via learning-to-think methodologies. 
The proposed approach is built upon two key insights: 
1) Long-CoT reasoning, i.e., inference-time scaling, has the potential to improve the accuracy of RM, especially in reasoning scenarios.
2) Taking the judging process as a verifiable task, we can further optimize the generative reward model by rejection sampling and reinforcement learning, similar to LLMs.
Based on the proposed framework, we develop \model series, including \modelnospace-32B and \modelnospace-32B-MATH, a collection of generative reward models with deep thinking abilities. Extensive results demonstrate that our \model series achieves state-of-the-art performance on various RM benchmarks, especially on reasoning-oriented benchmarks such as \benchnospace. 

We further conduct comprehensive RL experiments to analyze our \bench and \modelnospace. The experimental results demonstrate the correlation between our \bench
and downstream application, and the potential of Libra-RM in further reasoning data scaling with unlabeled data.

To summarize, our main contributions are as follows:
\begin{itemize}[leftmargin=*]
    \item We curate a reasoning-oriented RM benchmark from a diverse collection of challenging mathematical problems and
    advanced reasoning models, named \benchnospace, to address the limitations of existing RM benchmarks in reasoning scenarios.
    \item We propose a novel approach to improve generative reward model via learning to think, which yields \model series, a collection of powerful reward models that achieve state-of-the-art results on various RM benchmarks. 
    \item  Our RL experiments demonstrate a strong correlation between performance on Libra Bench and downstream application, as well as the potential of our \model in RL data scaling with unlabeled data.
\end{itemize}


\section{Related Work}\label{section:related_work}
\paragraph{Reward Models} 
Reward models (RMs) are designed to assign reward scores to responses generated by LLMs, and have been widely adopted in reinforcement learning, data selection, model evaluation, and other applications~\citep{zheng2023secrets, dong2024rlhf, zheng2023judging, li2024crowdsourced, dubois2024length,gu2024survey, seed2025seed}.
RMs are predominantly categorized into discriminative and generative types.
Discriminative reward models typically consist of an LLM backbone coupled with a value head.
They are trained on preference data with a classification objective and assign scalar rewards to responses~\citep{liu2024skywork, adler2024nemotron, wang2024interpretable}.
In contrast, generative reward models share the same architecture as standard LLMs but output textual judgments containing reward information for input responses~\citep{zhang2024generative,wang2024self, zhu2023judgelm, ankner2024critique, liu2025inference}. 
Notably, several works have proposed enhancing generative reward models with deep thinking capacities~\citep{chen2025judgelrm, chen2025rm, whitehouse2025j1incentivizingthinkingllmasajudge, guo2025rewardreasoningmodel}. However, fully leveraging the advantages of inference-time scaling for reasoning tasks and realizing the potential of thinking-enhanced generative reward models in downstream applications remain significant challenges.

\paragraph{Reward Model Benchmarks} 
Reward model benchmarks play a crucial role in guiding RM optimization and forecasting their performance on downstream applications~\citep{malik2025rewardbench, frick2024evaluate}. 
Conventional RM benchmarks predominantly target general question-answering tasks, assessing a model's ability to select the superior response in a pairwise setting~\citep{lambert2024rewardbench,frick2024evaluate,zhou2024rmb,tan2024judgebench, liu2024rm, saha2025learning}, which aligns with the Bradley-Terry (BT) model commonly employed in RM training~\citep{bradley1952rank}. 
This pairwise accuracy evaluation paradigm has been extended to other specific domains, including multimodal contexts, multilingual tasks, agentic systems, and more~\citep{gureja2024m, lu2025agentrewardbench, jin2024rag, wu2025rewordbench,chen2024mj,yasunaga2025multimodal,li2025vl,ruan2025vlrmbench}.
Recently, reasoning-oriented RM benchmarks have been proposed to evaluate the accuracies of (process) reward models in reasoning tasks~\citep{acemath2024, zheng2024processbench, song2025prmbench}.
However, these existing RM benchmarks suffer from one or two major limitations: the absence of challenging questions and responses from advanced
reasoning models, rendering them insufficient for assessing reward model in reasoning scenarios. 

\paragraph{Reinforcement Learning for LLMs} 
Reinforcement learning (RL) is widely employed in the post-training stage to enhance reasoning capabilities and align models with human preferences~\citep{ouyang2022training, guo2025deepseek}. 
Algorithms such as PPO, GRPO and their variants are predominantly used in RL for LLMs~\citep{schulman2017proximal, shao2024deepseekmath,yu2025dapo, yuan2025vapo}, while offline methods like DPO and KTO have also been proposed to accommodate resource-constrained environments~\citep{rafailov2023direct, ethayarajh2024kto}. 
RL for LLMs can be further classified by reward source into Reinforcement Learning from Human Feedback (RLHF) and Reinforcement Learning from Verifiable Reward (RLVR). 
RLHF leverages a reward model trained on human preference data to provide reward signals~\citep{ouyang2022training, bai2022training, zheng2023secrets, xiong2023iterative}, while RLVR optimizes models on verifiable tasks and receives rewards from rule-based answer matching and other predefined scripts~\citep{dong2024self, lambert2024t, guo2025deepseek}. 
In practice, RLHF and RLVR are often integrated to jointly optimize model behavior and mitigate the risk of reward hacking~\citep{liu2024deepseek, yang2025qwen3, seed2025seed}.

\section{Libra Bench}\label{sec:benchmark}

\begin{figure*}[t]
    \centering \includegraphics[width=1.0\columnwidth]{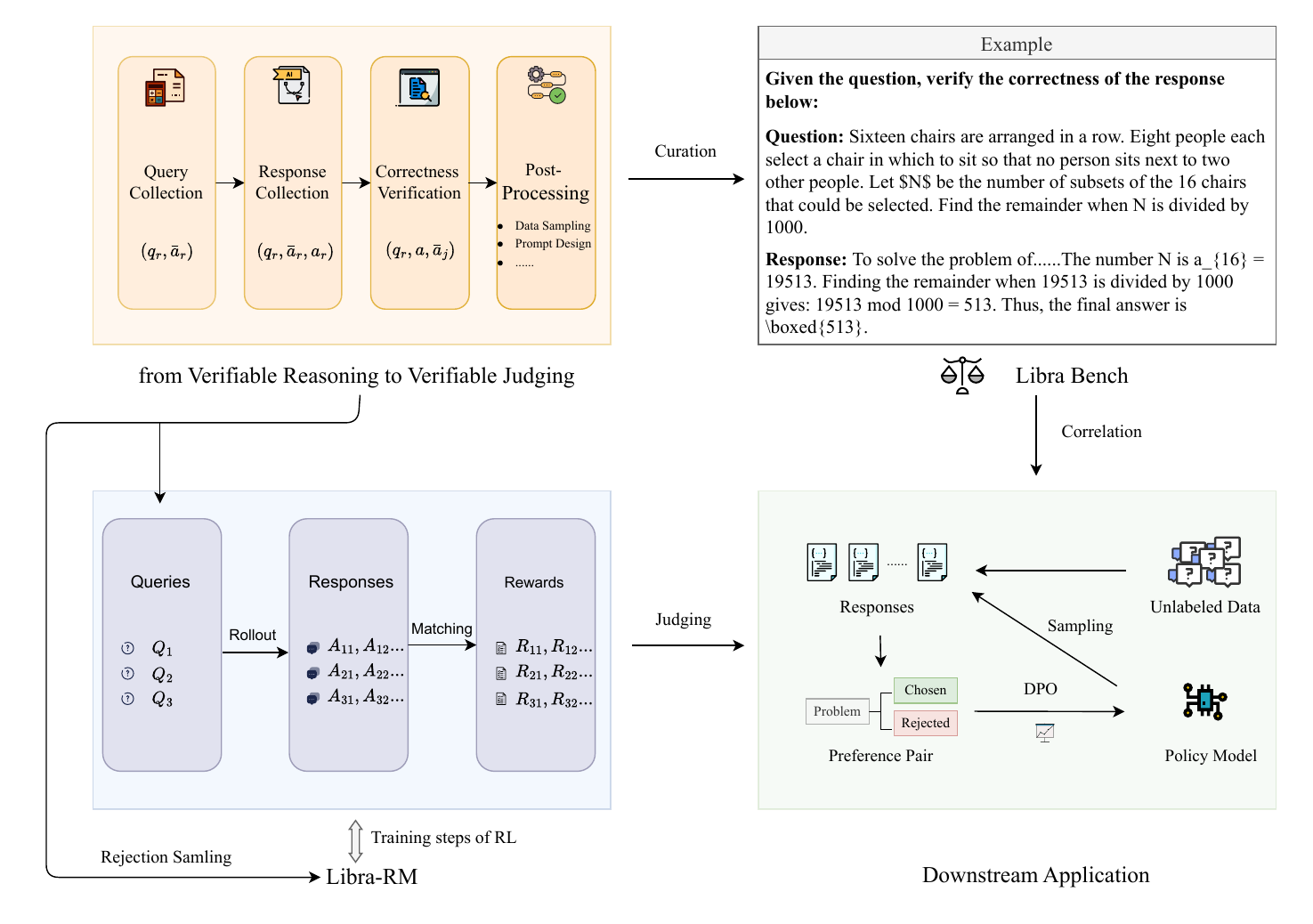}
    \caption{The overview of building \bench and \model.
    For \bench, we design the data strategy \textit{from Verifiable Reasoning to Verifiable Judging}, to curate RM benchmark from a collection of challenging mathematical problems and advanced reasoning models. 
    For \modelnospace, we adopt the same data strategy and combine reinforcement learning (RL) and rejection sampling for training.
    }
    \label{fig:libra_overview}
\end{figure*}
In this section, we detail the curation pipeline for \bench (Figure~\ref{fig:libra_overview}) and present its primary statistics and analysis.
Distinct from existing RM benchmarks, our \bench is constructed from a diverse set of challenging mathematical problems and advanced reasoning models, and is designed to assess pointwise accuracy in terms of correctness. 
These attributes ensure that \bench is well aligned with contemporary research, where reasoning models are primarily assessed and optimized for correctness on complex reasoning tasks.

\subsection{Pipeline: from Verifiable Reasoning to Verifiable Judging}\label{sec:benchmark:pipeline}
As illustrated in Figure~\ref{fig:libra_overview}, we curate the \bench with the strategy: from \textbf{V}erifiable reasoning to \textbf{V}erifiable judging (V2V), for RM evaluation. 
The total curation process consists of four stages: query collection, response collection, correctness verification and post-processing.
\paragraph{Query Collection} 
The query collection serves as the starting point of the entire curation pipeline. To adapt to the development of reasoning models, we collect 204 challenging mathematical problems from MATH-500 level5 (\cite{lightman2023let}), AIME 2024, and AIME 2025. Each problem is paired with a golden reference answer, covering various formats including integers, fractions, and formulas. Formally, each verifiable reasoning instance is denoted as $(q_r, \bar{a}_r)$, where $q_r$ is the reasoning problem and $\bar{a}_r$ is its golden reference answer.

\paragraph{Response Collection} Compared with existing RM benchmarks, we rollout generations from a collection of advanced reasoning models to assess the capacity of RM in complex reasoning tasks, including DeepSeek-R1~\citep{guo2025deepseek}, Qwen3-32B~\citep{yang2025qwen3}, QwQ-32B~\citep{qwen2025qwq}, DeepSeek-R1-Distill-Qwen-7B and DeepSeek-R1-Distill-Qwen-1.5B ~\citep{guo2025deepseek}. These models exhibit a wide range of accuracies (28.9\% - 81.4\% on AIME 2024), ensuring the diversity of our \benchnospace. For each problem $q_r$, we sample at least 64 responses from each model to guarantee a sufficient number of both correct and incorrect replies. At this stage, each data point is formulated as  $(q_r, \bar{a}_r, a_r)$, where $a_r$ is the sampled response for $q_r$.

\paragraph{Correctness Verification} 
We annotate the outcome correctness of each response $a_r$ based on the problem $q_r$ and the reference answer $\bar{a}_r$, thereby transforming reasoning problems to judging problems.
In practice, we employ a sophisticated combination of different methods to ensure the reliability of correctness verification, including rule-based answer matching, model-based evaluation and human annotation. 
Notably, our model-based evaluation leverages advanced reasoning models to annotate responses against golden references, enabling robust handling of complex answer formats with high accuracy, akin to~\citet{seed2025seed}.
Details of the methodologies and statistics of our annotation are reported in Appendix~\ref{ap: bench: veri}. 

We denote the label of correctness as $\bar{a}_j$ which takes binary values 0 or 1. Each sample is thus represented as $(q_r, a_r, \bar{a}_j)$, where the $\bar{a}_r$ is omitted after annotation as in existing RM benchmarks. 
The change of subscript from $r$ to $j$ indicates a transition from \textbf{r}easoning problem to a \textbf{j}udging problem. The correctness label $\bar{a}_j$ serves as the reference answer of judging problem $q_j$, which is derived from the concatenation of $q_r$, $a_r$, and a predefined prompt template. 

\paragraph{Post-processing} We further perform several post-processing steps to refine our \benchnospace. 
First, We remove the Chain-of-Thought (CoT) segments from sampled responses, as they often involve complex trial-and-error processes that are not supervised in mainstream training paradigms~\citep{guo2025deepseek}. The truncated samples containing only the CoT component are also filtered out.
Secondly, we balance the proportion of our \bench such that each model contributes an equal number of correct and incorrect responses in each data subset, as detailed in Table~\ref{tab:bench:libra_stat}.

For evaluation, the RM receives the concatenation of $q_r$, $a_r$, and the predefined prompt template as input, determines the correctness of $a_r$, and outputs a binary prediction. 
Benefiting from the balanced distribution, we directly calculate the accuracies across different data subsets to assess the capacity of RM in reasoning scenarios, as shown in Table~\ref{tabel: main_librabench}. 

\begin{table}[!t]
  \centering
  \small
  \vspace{-1mm}
  \scalebox{0.95}{
    \begin{tabular}{lcccccc}
    \toprule
      & \multicolumn{2}{c}{\textbf{MATH}} & \multicolumn{2}{c}{\textbf{AIME 2024}} & \multicolumn{2}{c}{\textbf{AIME 2025}} \\
    \cmidrule(lr){2-3}\cmidrule(lr){4-5}\cmidrule(lr){6-7}
    & \textbf{correct} & \textbf{incorrect} & \textbf{correct} & \textbf{incorrect} & \textbf{correct} & \textbf{incorrect} \\
    \midrule
    \multirow{1}[0]{*}{\textbf{Number of Problems}} & 134 & 134 & 30 & 30  & 30   & 30   \\
    \midrule
    \multirow{1}[0]{*}{\textbf{Number of Samples}} & 680   & 680 & 600 & 600 & 600   & 600  \\
   
    \multirow{1}[0]{*}{\ \ \ \ DeepSeek-R1} & 134   & 134   & 120   & 120   & 120   & 120   \\
    \multirow{1}[0]{*}{\ \ \ \ Qwen3-32B} & 134   & 134   & 120   & 120   & 120   & 120   \\
    \multirow{1}[0]{*}{\ \ \ \ QwQ-32B} & 134   & 134   & 120   & 120   & 120   & 120   \\
    \multirow{1}[0]{*}{\ \ \ \ R1-Distill-Qwen-7B} & 134   & 134   & 120   & 120   & 120   & 120   \\
    \multirow{1}[0]{*}{\ \ \ \ R1-Distill-Qwen-1.5B} & 134   & 134   & 120   & 120   & 120   & 120   \\
    \midrule
    \multirow{1}[0]{*}{\textbf{Annotation Approach}} & \multicolumn{2}{c}{Model-based + Human}   & \multicolumn{2}{c}{Rule-Based}   & \multicolumn{2}{c}{Rule-Based}   \\

    \bottomrule
    \end{tabular}%
  }
  \vspace{2mm}
  \caption{Statistics of the \benchnospace. The \bench comprises problems sourced from MATH-500 level5, AIME 2024, AIME 2025, with responses generated from Deepseek-R1, Qwen3-32B, QwQ-32B, DeepSeek-R1-Distill-Qwen-7B, and DeepSeek-R1-Distill-Qwen-1.5B. 
  We divide the \bench into three subsets based on the problem sources. Samples derived from the MATH-500 level5 are annotated by model-based evaluation and human annotation, while those derived from  the AIME 2024 and AIME 2025 are annotated via rule-based answer matching.}
  \label{tab:bench:libra_stat}
\end{table}%

\subsection{Statistics and Analysis}\label{sec:benchmark:stat}
We present the basic statistics of our \bench in Table~\ref{tab:bench:libra_stat}. 
Our \bench consists of 3,740 samples which are curated from 204 challenging mathematical problems and 5 advanced reasoning models. 
More examples of our \bench can be found in Appendix~\ref{ap: bench: example}.

We further evaluate state-of-the-art reward models and LLM-as-a-Judge methods in our \benchnospace, and report their performance in Table~\ref{tabel: main_librabench}. 
For discriminative RM, we greedily search for a threshold to binarize scalar scores and maximize the average accuracy. For generative RM and LLM-as-a-Judge, we experiment with various prompt templates and report the best accuracy. 
Compared with the reasoning subsets of existing RM benchmarks, most models achieve lower accuracy on our \bench, owing to both the increased difficulty of the problems and the presence of confusing responses from advanced reasoning models. 
From Table~\ref{tabel: main_librabench}, we can also observe the superior performance of thinking models over non-thinking models on our \bench, with non-thinking models achieving 55.1\%-69.1\% accuracy and thinking models achieving 73.7\%-78.7\% (excluding our \modelnospace). These findings motivate further improvements in RM accuracy via learning-to-think methodologies, which will be discussed in Section~\ref{section:apporach}.

\section{Approach for Libra-RM}\label{section:apporach}
To overcome the limitations of current RMs in reasoning scenarios, we propose a comprehensive approach to improve generative reward models via learning-to-think, resulting in our \model series. 
As illustrated in Figure~\ref{fig:libra_overview}, our \model series are trained through a combination of rejection sampling and reinforcement learning, sharing the same data strategy as our \benchnospace: from \textbf{V}erifiable reasoning to \textbf{V}erifiable Judging (V2V). 

In this section, we formulate our task definition in subsection~\ref{section:apporach:pre}, followed by the details of our rejection sampling and reinforcement learning in \ref{section:method:rs} and \ref{section:method:rl}, respectively.

\subsection{Task Definition} \label{section:apporach:pre}
Conventional reward models (RMs) are typically discriminative, mapping an input query and a candidate response to a scalar quality score. In contrast, we explore a generative paradigm for reward modeling. A generative RM is a text-to-text model conditioned on a query $q$, a set of candidate responses $[a_1, a_2, \ldots]$, and a set of evaluation criteria $c$. Its objective is to generate a natural language judgment $j$ that evaluates the candidate answers according to the specified criteria. Formally, this process is defined as:
\begin{equation}
 j = \mathrm{RM}_{\mathrm{gen}}(q, [a_1, a_2, \ldots], c) 
\end{equation}
Based on the criteria $c$, the judging tasks are further categorized into scoring and ranking. 
In the scoring setting, the generative reward model is required to assign a specific rating $\text{score}_i$ to each answer $a_i$. In contrast, the ranking task only requires the model to define the relative $\text{rank}_i$ for $a_i$ within the answer list. 
Both the scores and rankings can be extracted from the textual judgment $j$.

\begin{equation}
\text{RM}(q, [a_1, \ldots], c) = 
\begin{cases}
[\text{score}_1,\ldots] &  c \in \mathbb{C}_{score} \\
[\text{rank}_1, \ldots] &  c \in \mathbb{C}_{rank}
\end{cases}
\end{equation}
In this work, we develop \modelnospace-32B-MATH and \modelnospace-32B, both endowed with advanced deep thinking capabilities. 
The \modelnospace-32B-MATH is specialized for the reasoning-oriented pointwise scoring task in terms of correctness, achieving state-of-the-art performance on \bench and downstream applications. 
The \modelnospace-32B is an extended version, which is also capable of preference ranking tasks as existing RMs, further demonstrating the generalizability of our approach.

\subsection{Rejection Sampling and Supervised Fine-Tuning}\label{section:method:rs}
We perform rejection sampling based on DeepSeek-R1 and finetune our \model from the pretrained model Qwen2.5-32B to accelerate convergence and improve accuracy. 
During this phase, both judging and non-judging data are collected to enhance the diversity of the training dataset and boost the performance of our \modelnospace.

\paragraph{Judging data}
For pointwise scoring, we initially collect a set of labeled verifiable judging data in reasoning scenarios via the V2V strategy. 
The correctness labels are annotated through model-based evaluation, employing advanced reasoning models to verify the answer correctness against the golden reference answer.
We then perform rejection sampling on DeepSeek-R1 to collect responses that are consistent with $\bar{a}_j$, forming the subset $\mathcal{D}^{rs}_{score}$.
For pairwise ranking, we directly utilize existing preference data for RM training. The input prompt is formed by the concatenation of the problem $q$, answer pair $[a_1, a_2]$ and the judgment criteria $c_{rank}$. The golden references for judgment are taken from the original annotation in preference data.
We apply the same rejection sampling procedure as in the pointwise scoring data to construct the data subset $\mathcal{D}^{rs}_{rank}$.

\paragraph{Non-Judging data}
The curation of non-judging data adheres to the standard SFT setting, with prompts comprising only the question and ground truths as direct answers. 
For reasoning tasks, we perform rejection sampling on verifiable reasoning problems. And for general tasks (non-reasoning), we directly sample generations from DeepSeek-R1 as the ground truth without rejection, thereby ensuring the entire training process is RM-free. In this way, we obtain two data subsets $\mathcal{D}^{rs}_{reason}$ and $\mathcal{D}^{rs}_{general}$.
In practice, we combine the $\mathcal{D}^{rs}_{score}$ and $\mathcal{D}^{rs}_{reason}$ subsets to train \modelnospace-32B-Math, while all data subsets are utilized for training \model-32B. 
An ablation study on our data composition is presented in Section~\ref{sec:exp:ablation}. 

\subsection{reinforcement Learning for judging}\label{section:method:rl}
Following rejection sampling and supervised fine-tuning, we further apply rule-based reinforcement learning on a verifiable dataset to improve the accuracy of our \modelnospace.
We detail our training recipe from three aspects: data, reward design, and learning objective.

\paragraph{Data} 
Similar to the rejection sampling, we curate a mixed RL dataset consisting of judging data and non-judging data for training.
The judging data consists of $\mathcal{D}^{rl}_{score}$ for pointwise scoring and $\mathcal{D}^{rl}_{rank}$ for pairwise ranking, while the non-judging data consists solely of verifiable reasoning data $\mathcal{D}^{rl}_{reason}$. 
All of our RL dataset are verifiable, and the entire training process does not depend on any other RMs. 
In practice, we combine $\mathcal{D}^{rl}_{score}$ and $\mathcal{D}^{rl}_{reason}$ subsets for training \modelnospace-32B-MATH, while all three subsets are incorporated for training \modelnospace-32B.

\paragraph{Reward Design}
We adopt a rule-based reward signal that consists of correctness reward and length penalty for training \modelnospace, formulated as  Equation~\ref{eq:all_reward}. 
\begin{equation} \label{eq:all_reward}
\text{reward}(x, y, \bar{y}) =  \text{is}\_\text{correct}(y,\bar{y}) - \text{len}\_\text{penalty}(y)
\end{equation}
where $x$ denotes the input, $y$ denotes the response, and $\bar{y}$ denotes the ground-truth answer for $x$.

The correctness reward $\text{is}\_\text{correct}(y,\bar{y})$ is computed via rule-based answer matching to assess the correctness of the response's final outcome. 
Typically, the final outcome of the response can be extracted as a boxed number in reasoning tasks or as a formatted verdict in judging tasks. 
If the extracted outcome aligns with the golden reference answer, a correctness reward of 1 is assigned; otherwise, a reward of 0 is given.

We also incorporate a length penalty into our reward system, which has been demonstrated effective in performance improvement and length compression~\citep{yu2025dapo, team2025kimi}. 
As formulated in Equation~\ref{eq:len_penalty}, the length penalty is defined as the ratio of the excess length over the expected length to the buffer length, where $L_{exp}$ denotes the expected length and $L_{max}$ represents the maximum generated length during training.
The buffer length is given by the difference $L_{max} - L_{exp}$ and length penalty is constrained to the range $[0, 1]$ since responses longer than $L_{max}$ are truncated.
\begin{equation} \label{eq:len_penalty}
\text{len}\_\text{penalty}(y) = \text{max}(\frac{ |y| - L_{exp}}{L_{max} - L_{exp}}, 0)
\end{equation}

\paragraph{Learning Objective} 
We adopt GRPO~\citep{shao2024deepseekmath} with Clip-Higher strategy~\citep{seed2025seed} as our reinforcement learning algorithm motivated by its resource efficiency and strong empirical performance. The learning objective can be formulated as follows:
\begin{equation} \small
\begin{aligned}
J_{GRPO}(\theta) &= \mathbb{E}[q \sim P(Q), \{o_i\}^G_{i=1} \sim \pi_{\theta_{old}}(O|q)] \\
& \frac{1}{G} \sum^{G}_{i=1} \frac{1}{|o_i|} \sum^{|o_i|}_{t=1} \left( \min \left( r_{i,t}(\theta) a_{i,t}, \text{clip} \left( r_{i,t}(\theta), 1-\varepsilon_{low}, 1+\varepsilon_{high} \right)a_{i,t} \right) - \beta D_{KL} \left( \pi_{\theta}||\pi_{ref} \right) \right)
\end{aligned}
\end{equation}
where $r_{i,t}(\theta) = \frac{\pi_{\theta}(o_{i,t}|q, o_{<t})}{\pi_{\theta_{old}}(o_{i,t}|q, o_{<t})}$, and  $D_{KL}$ denotes an unbiased estimator of the KL divergence, formulated as
$D_{KL} \left( \pi_{\theta}||\pi_{ref} \right) = \frac{\pi_{ref}(o_i|q)}{\pi_{\theta}(o_i|q)} - \log \frac{\pi_{ref}(o_i|q)}{\pi_{\theta}(o_i|q)} - 1 $. The advantage $a_{i,t}$ represents the relative reward of the output within the corresponding group, calculated as
$a_{i,t} = \frac{r_i - \text{mean}(\{r_1, r_2, \cdots, r_G\})}{\text{std}(\{r_1, r_2, \cdots, r_G\})}$.

\section{Experiments on RM Benchmarks}\label{sec:exp}
We conduct extensive experiments to evaluate and analyze both our \bench and our \modelnospace. 
We begin by detailing our experimental setups in subsection~\ref{sec:exp:setup}.
Subsequently, we present the performance of \model and various baseline methods on \bench in subsection~\ref{sec:exp:libra_bench}.  
Furthermore, subsection~\ref{sec:exp:rm_bench} extends the evaluation to other widely adopted RM benchmarks, highlighting the generalizability of our approach. 

\subsection{Experimental Setups} \label{sec:exp:setup}
\paragraph{Benchmarks} We evaluate our \model and baseline models on both our proposed \bench and existing RM benchmarks. 
\bench, detailed in section~\ref{sec:benchmark}, is specially designed to assess the pointwise accuracy of RMs on challenging reasoning tasks.
To ensure a comprehensive comparison, we also include widely used RM benchmarks such as Reward Bench, PPE Preference, PPE Correctness, RMB, and JudgeBench \citep{lambert2024rewardbench,frick2024evaluate,zhou2024rmb,tan2024judgebench}, which are designed to measure the pairwise accuracy of RMs in general scenarios.

\begin{table*}[ht] 
\centering
\small
\begin{tabular}{lcccc}
\toprule
Model & MATH-500 & AIME2024 & AIME2025 & Average \\
\midrule
\midrule
\multicolumn{4}{c}{Discriminative Reward Models} \\
InternLM2-20B-Reward & 59.9 & 67.1& 62.2 &63.1 \\
Skywork-Reward-Gemma-2-27B  & 55.8  &54.5 & 55.1&55.1  \\ 
ArmoRM-8B-v0.1  & 57.2 & 61.8 & 58.9&59.3 \\ 
Qwen2.5-Math-RM-72B  & 69.9 & 69.1 & 58.0 &65.7 \\
AceMath-72B-RM  & 73.6 & 65.4 & 60.8 & 66.6 \\

\midrule
\multicolumn{4}{c}{LLM-as-a-Judge} \\
GPT-4o-0816 & 69.9 & 66.1  & 61.4 & 65.8 \\
GPT-4.1 & 71.3 & 71.0  & 65.0  & 69.1  \\
Claude-3.5-sonnet  & 64.9 & 65.2 & 63.9 & 64.7 \\
Claude-3.7-sonnet  &  70.8 & 65.6 & 65.0 & 67.1 \\
Llama-3.1-70B-Instruct & 50.8  & 50.4 & 51.4 & 50.9\\

\midrule
\multicolumn{4}{c}{LLM-as-a-Judge with thinking} \\
DeepSeek-R1 & 82.2 & 76.8 & 77.4 & 78.8 \\
Qwen3-32B & 80.2  & 78.3 & 75.8 & 78.1  \\
QwQ-32B & 80.8 & 77.1 & 74.7 & 77.5\\
R1-Distill-Qwen-32B & 75.9 & 75.0 &  70.2 & 73.7 \\

\midrule
\multicolumn{4}{c}{Generative Reward Models} \\
Skywork-Critic-Llama-3.1-70B & 55.4 & 60.6 & 57.2 & 57.7 \\ 
\modelnospace-32B-MATH (Ours) & \textbf{83.4} & \textbf{81.5}  & \textbf{80.3} & \textbf{81.7}\\
\modelnospace-32B (Ours) &  \underline{82.8} & \underline{79.7} & \underline{77.5} & \underline{80.0}\\
\bottomrule
\end{tabular}

\caption{Evaluations on \bench. Bold numbers indicate the best performance, while underlined numbers indicate the second-best performance among baseline models and our \modelnospace. For generative reward model, the best accuracy across different prompt templates is reported (see Appendix~\ref{sec: ap: prompts}). For discriminative reward model, we select a threshold that maximizes the average accuracy to convert model outputs into binary correctness predictions. Note that all metrics reported in the table are accuracies and the subsets MATH-500, AIME2024, and AIME2025 only refer to the sources of problems.}
\label{tabel: main_librabench}
\end{table*}

\paragraph{Baseline methods} 
We compare our \model with leading reward models and LLM-as-a-Judge methods:
\begin{itemize}[leftmargin=*]
    \item Discriminative Reward Models: InternLM2-20B-Reward ~\citep{cai2024internlm2}, Skywork-Reward-Gemma-2-27B~\citep{liu2024skywork}, ArmoRM-8B-v0.1~\citep{ArmoRM}, Nemotron-4-340B-Reward~\citep{adler2024nemotron}, Qwen2.5-Math-RM-72B~\citep{yang2024qwen25mathtechnicalreportmathematical}, AceMath-72B-RM~\citep{acemath2024}.
    \item Generative Reward Models: Skywork-Critic-Llama-3.1-70B\citep{skyworkcritic2024}, DeepSeek-GRM-27B\citep{liu2025inference}.
    \item LLM-as-a-Judge methods: GPT-4o-0816~\citep{hurst2024gpt}, GPT-4.1~\citep{gpt41}, Claude-3.5-sonnet~\citep{claude35sonnet}, Claude-3.7-sonnet~\citep{claude37sonnet}, Gemini-1.5-pro~\citep{team2024gemini}, Llama-3.1-70B-Instruct~\citep{grattafiori2024llama}.
    \item LLM-as-a-Judge methods with thinking: DeepSeek-R1~\citep{guo2025deepseek}, Qwen3-32B~\citep{yang2025qwen3}, QwQ-32B~\citep{qwen2025qwq}, R1-Distill-Qwen-32B~\citep{guo2025deepseek}.
    \item \model Series: Our proposed generative Reward Models with deep thinking capabilities. \model series were trained through the approach in ~\ref{section:apporach}, and training details are reported in ~\ref{ap: train}.
\end{itemize}

\subsection{Evaluations on our \bench} \label{sec:exp:libra_bench}
We first compare the performance of our \model with various baseline models on the \benchnospace. 
As shown in Table \ref{tabel: main_librabench}, our \modelnospace-32B-MATH and \modelnospace-32B consistently outperform all baselines across all subsets of \benchnospace. 
Specifically, our \modelnospace-32B-MATH attains an average accuracy of 81.7, while AceMath-72B-RM attains the highest accuracy of 66.6 among discriminative reward models and GPT-4.1 attains the highest accuracy of 69.1 among LLM-as-a-Judge methods without deep thinking. 
We also compare our \model with state-of-the-art thinking models.
Trained from the same base model, our \modelnospace-32B-MATH outperforms the QwQ-32B and R1-Distill-Qwen-32B by significant margins (81.7 vs. 77.5 and 81.7 vs. 73.7, respectively), demonstrating the effectiveness of our proposed approach.
Notably, \modelnospace-32B-MATH even surpasses Qwen3-32B and DeepSeek R1, which were trained from stronger base models, achieving accuracy gains of 3.6 and 2.9, respectively. 

To further understand these performance differences, we analyze the confusion matrices on \bench, as shown in Table~\ref{tabel: confusion}. 
The results indicate that verifying incorrect samples is substantially more challenging than verifying correct ones.
Notably, our \model series demonstrates superior performance in handling incorrect samples and achieves the highest macro F1 score among all evaluated models.

\begin{table*}[t] 
\centering
\small
\begin{tabular}{lccccc}
\toprule
Model & TP & TN & FP & FN & Macro F1\\
\midrule
\midrule
GPT-4.1 & 1281 & 1307 & 563 & 589 & 0.692 \\
Claude-3.7-sonnet & 1448 & 1068 & 802 & 422 & 0.669 \\
DeepSeek-R1 & 1693 & 1258 & 612 & 177 & 0.786 \\
Qwen3-32B & 1652  & 1272 & 598 & 218 & 0.780 \\
QwQ-32B & 1670 & 1234 & 636 & 200 & 0.773 \\
R1-Distill-Qwen-32B & 1689 & 1070 &  800 & 181 & 0.730 \\
\modelnospace-32B-MATH (Ours) & 1662 & 1397  & 473 & 208 & 0.817 \\
\modelnospace-32B (Ours) &  1601 & 1395 & 475 & 269 & 0.800\\
\bottomrule
\end{tabular}
\caption{Confusion matrices on \benchnospace. TP, TN, FP, and FN are short for True Positive, True Negative, False Positive, and False Negative. 
Macro F1 is calculated as the arithmetic mean of the F1 scores for positive samples and negative samples.}
\label{tabel: confusion}
\end{table*}

\subsection{Evaluations on general RM benchmarks} \label{sec:exp:rm_bench}
We further evaluate our \model with existing reward models and LLM-as-a-Judge methods on widely used RM benchmarks for comprehensive assessment.
Table~\ref{tabel: main_chat} reports the overall scores of our \model and baseline models on various RM benchmarks, including Reward Bench, PPE Preference, PPE correctness, RMB, and JudgeBench \citep{lambert2024rewardbench,frick2024evaluate,zhou2024rmb,tan2024judgebench}.  
Different from \benchnospace, these RM benchmarks require RMs to predict the preference between two responses.
As shown in Table~\ref{tabel: main_chat}, our \modelnospace-32B outperforms both existing reward models and LLM-as-a-Judge methods in terms of average accuracy.
Specifically, \modelnospace-32B attains the PPE correctness accuracy of 77.3 and the JudgeBench accuracy of 77.1, substantially surpassing the baseline models. As for other RM benchmarks such as RewardBench, PPE preference, and RMB, our \modelnospace-32B still achieves competitive results, consistently ranking among the top tier. 
Table~\ref{tab: app: reason} further demonstrates the advantages of our \model on the reasoning subsets of these RM benchmarks. 
Notably, compared with existing reward models, \modelnospace-32B exhibits strong stability and delivers outstanding performance on all RM benchmarks.
Detailed scores of our model and the baseline models are presented in Appendix~\ref{ap:app:eval}. 

\begin{table*}[t] 
\centering

\begin{adjustbox}{width=1\columnwidth,center}
    \begin{tabular}{lcccccc}
    \toprule
    Model & Reward Bench & PPE-P & PPE-C & RMB & JudgeBench & Average \\
    \midrule
    \midrule
    \multicolumn{6}{c}{Discriminative Reward Model} \\
    InternLM2-20B-Reward & 90.2 & 61.0 & 63.0& 62.9 & 63.4 &68.1  \\
    Skywork-Reward-Gemma-2-27B  & 94.3 &56.6 &56.6 &60.2 &64.3 & 66.4 \\ 
    ArmoRM-8B-v0.1  & 90.4 &60.6 &61.2 &64.6 & 56.9 & 66.7\\ 
    Nemotron-4-340B-Reward & 92.0 & 59.3 & 60.8 & 69.9 & - & -\\
    
    \midrule    
    \multicolumn{6}{c}{LLM-as-a-Judge} \\
    GPT-4o-0816  & 86.7 & 67.7  & - & - & 56.6 & - \\
    Claude-3.5-sonnet   & 84.2 & 67.3  & 68.4 & 70.6 & 64.3 & 71.0 \\
    Gemini-1.5-pro  & 86.8 & 66.1  & 59.8 & 56.5 & 47.1 &63.3 \\
    Llama-3.1-70B-Instruct  & 84.0& 65.3& 63.2 & 68.9&52.3 & 66.7\\
    \midrule    
    \multicolumn{6}{c}{Generative Reward Model} \\
    DeepSeek-GRM-27B  & 86.0 & 64.7 & 59.8 & 69.0 & - & - \\
    \modelnospace-32B-MATH (Ours) & 89.1  & 63.9 & 75.2& 65.5 &76.6 & 74.1 \\
    \modelnospace-32B (Ours) & 92.9 & 66.5 & 77.3 & 72.9& 77.1 & 77.3 \\
    \bottomrule
    \end{tabular}
\end{adjustbox}
\caption{Overall evaluations on mainstream reward model (RM) benchmarks for pairwise ranking tasks. PPE-P is short for PPE preference, and PPE-C is short for PPE correctness. Bold numbers indicate the best performance, while underlined numbers indicate the second-best performance among all baseline and our \model . Baseline results are taken from previous work (\cite{liu2025inference, frick2024evaluate, zhou2024rmb}, with missing JudgeBench scores supplemented by us.}
\label{tabel: main_chat}
\end{table*}

\section{Experiments on Downstream Tasks}\label{sec:exp:downstream}
We further conduct a series of DPO experiments to investigate the relationship between the accuracy of Libra-Bench and the performance of downstream application. The experimental results also demonstrate the potential of our \model for RL data scaling.
\subsection{Experimental Setups}
We select the R1-Distill-Qwen-7B and R1-Distill-Llama-8B as the initial policy models to investigate how reward models of varying accuracies impact reasoning performance via DPO. 
The queries for DPO are drawn from Skywork-OR1-RL-Data~\citep{he2025skywork}, and for each initial policy model, and we sample 4 responses on each query for DPO training.
The DPO experiments are conducted with seven different reward models, spanning various categories and exhibiting different levels of accuracy on \benchnospace. 
We instruct the reward models to annotate the correctness of the sampled responses without access to reference answers, simulating the scenario of RL data scaling on unlabeled data.
The annotations are performed following the same evaluation protocol as in Table~\ref{tabel: main_librabench}, and each preference pair consists of one correct and one incorrect response as labeled by the reward models.
For each query, we curate preference pairs by matching the minimum number of correct and incorrect responses.

As for hyperparameters, the $\beta$ is set to 0.01, the global batch size is set to 256, the training epoch is set to 3, and the learning rate is set to $10^{-6}$. As for evaluation, the temperature is set to 0.6, and the maximum number of new tokens is set to 32,768. We calculate pass@1 scores on AIME 2024 and AIME 2025 by sampling 32 responses per query.

\begin{figure}[t]
  \centering
  \begin{minipage}[t]{0.38\textwidth}
    \centering
    \includegraphics[width=0.91\textwidth]{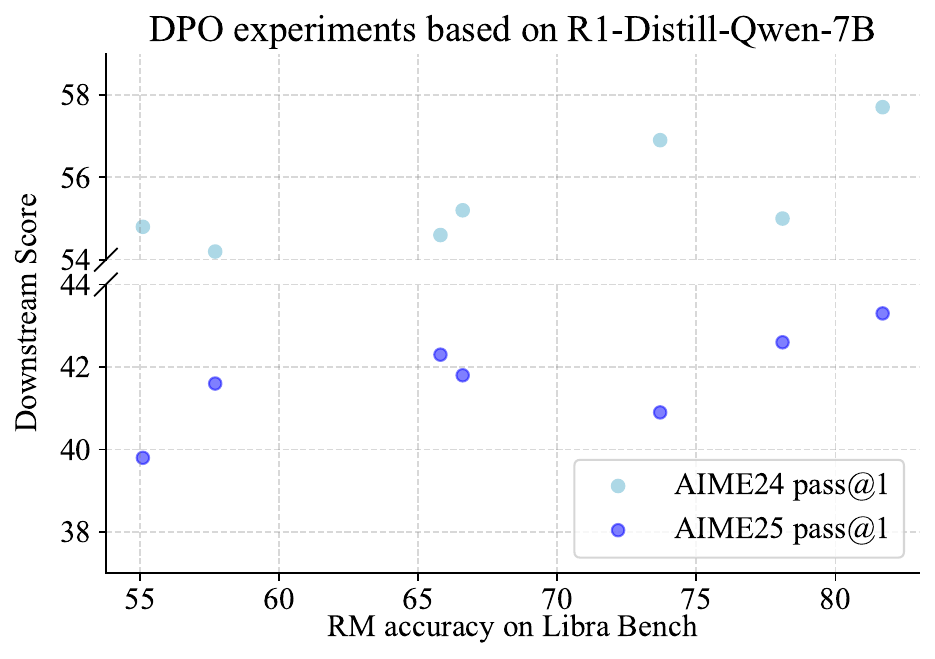}

    \label{fig:libra_aime_7b}

    \includegraphics[width=0.91\textwidth]{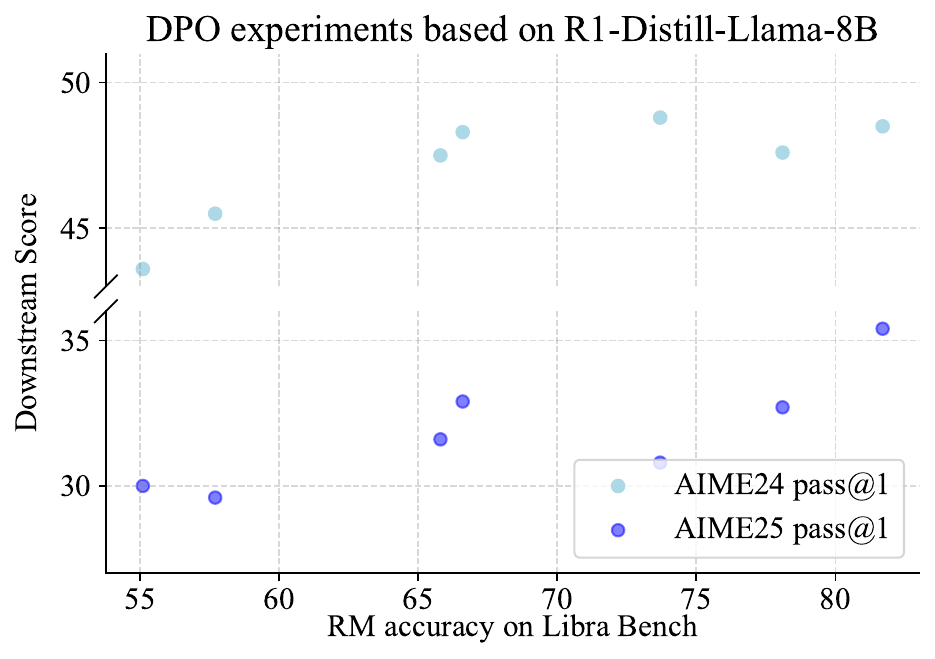}

    \label{fig:libra_aime_8b}
  \end{minipage}
  \hfill
  \begin{minipage}[t]{0.61\textwidth}
    \small
\begin{tabular}{lccc}
\toprule
Model  &AIME24&AIME25&RM ACC\\
\midrule
\midrule
R1-Distill-Qwen-7B$\dagger$ & 55.5 & 39.2 & - \\
 + DPO$\times$Skywork-RM  & 54.8 & 39.8 & 55.1 \\
 + DPO$\times$Skywork-Critic  & 54.2 & 41.6 & 57.7\\
 + DPO$\times$GPT-4o-0816  & 54.6 & 42.3 & 65.8 \\
 + DPO$\times$AceMath-RM  & 55.2 & 41.8  & 66.6\\
 + DPO$\times$R1-Qwen-32B  & 56.9 & 40.9 & 73.7 \\ 
 + DPO$\times$Qwen3-32B  & 55.0 & 42.6 & 78.1 \\ 
 + DPO$\times$Libra-MATH  & 57.7 & 43.3 & 81.7\\ 
\midrule    
\midrule
R1-Distill-Llama-8B$\ddagger$  & 43.1(50.4) & 30.7 & -\\
 + DPO$\times$Skywork-RM & 41.9 & 30.0 & 55.1 \\
  + DPO$\times$Skywork-Critic & 45.5 & 29.6 & 57.7 \\
  + DPO$\times$GPT-4o-0816 & 47.5 & 31.6 & 65.8 \\
  + DPO$\times$AceMath-RM & 48.3 & 32.9 & 66.6 \\ 
  + DPO$\times$R1-Qwen-32B & 48.8 & 30.8 & 73.7 \\ 
  + DPO$\times$Qwen3-32B & 47.6 & 32.7 & 78.1 \\
 + DPO$\times$Libra-MATH  & 48.5 & 35.4 & 81.7 \\
 
\bottomrule
\end{tabular}
  \end{minipage}
\caption{Correlation between \bench accuracy and downstream performance.  $\dagger$: Results taken from \citet{guo2025deepseek} and \citet{wen2025light}. $\ddagger$: We re-evaluate the metrics for R1-Distill-Llama-8B, with the results in parentheses taken from \citet{guo2025deepseek}.}
\label{fig:result_ensemble}
\end{figure}

\subsection{Results}
We illustrate the correlation between the accuracy of Libra-Bench and the performance of downstream DPO experiments in Figure~\ref{fig:result_ensemble}. For brevity, we use the following abbreviations: ``Skywork-RM''  refers to Skywork-Reward-Gemma-2-27B, ``Skywork-Critic'' to Skywork-Critic-Llama-3.1-70B, ``AceMath-RM'' to AceMath-72B-RM, ``R1-Qwen-32B'' to R1-Distill-Qwen-32B, and ``Libra-MATH'' to Libra-RM-32B-MATH, ``RM ACC'' to RM accuracy on \benchnospace.

For both R1-Distill-Qwen-7B and R1-Distill-Llama-8B, the RM accuracies on our \bench show a consistent correlation with downstream DPO performance, as measured by pass@1 scores on AIME 2024 and AIME 2025. 
Our results reveal that existing RMs and LLM-as-a-Judge methods are limited in enhancing reasoning performance through DPO, primarily due to their low accuracies. 
In contrast, models equipped with deep thinking capabilities consistently achieve higher accuracy on \bench and demonstrate superior downstream performance in our experiments. 
The correlation between \bench accuracy and downstream application performance highlights the utility of our \bench in guiding RM optimization and predicting RM performance.

Among these models, our \modelnospace-32B-MATH achieves the best performance on both the \bench evaluation and downstream DPO experiments. 
For initial policy model R1-Distill-Qwen-7B, our \modelnospace-32B-MATH improves the pass@1 score on AIME 2024 from 55.5\% to 57.7\%, and on AIME 2025 from 39.2\% to 43.3\%. 
Similarly, for initial policy model R1-Distill-Llama-8B, our \modelnospace-32B-MATH also increases the pass@1 score of AIME 2025 from 30.7\% to 35.4\%.
Notably, all these enhancements are achieved without access to golden reference answers, demonstrating the potential of our \model for RL data scaling on unlabeled data.
\section{Ablation Studies and Discussion}\label{sec:exp:ablation}
In this section, we present ablation studies to analyze the effectiveness of different components in our proposed approach. 

\subsection{Ablation study on Multi-Stage Training}
We first examine the impact of SFT and RL stages in training \model-32B-MATH. 
We use \model-32b-MATH as the basis for our studies and supplement with an RL-zero experiment, where we directly apply reinforcement learning to Qwen2.5-32B rather than the SFT checkpoint.
\paragraph{Experimental Setups} 
Following ~\citet{seed2025seed, he2025skywork}, we set the coefficient of the KL loss to 0. All other hyperparameters and the training dataset for the RL-zero experiment are the same as those for \model-32B-MATH, as detailed in Appendix~\ref{ap: train}.  
We specially adjust the prompt template to incentivize the model's thinking capacity, similar to the DAPO dataset~\citep{seed2025seed}.

\begin{figure}[t]
  \centering
  \begin{subfigure}[b]{0.49\textwidth}
    \includegraphics[width=\textwidth]{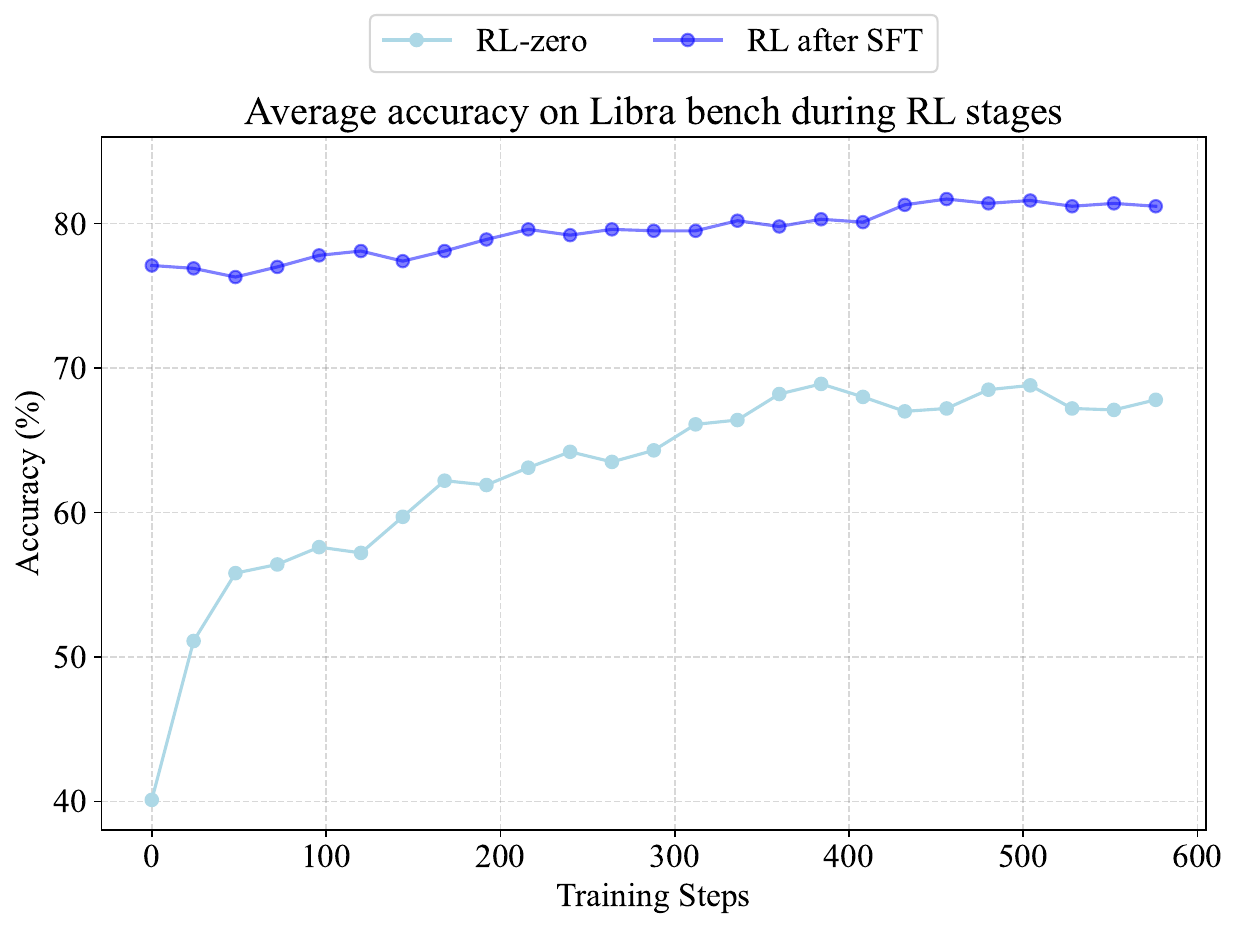}
    \label{fig:left}
  \end{subfigure}
  \hfill 
  \begin{subfigure}[b]{0.49\textwidth}
    \includegraphics[width=\textwidth]{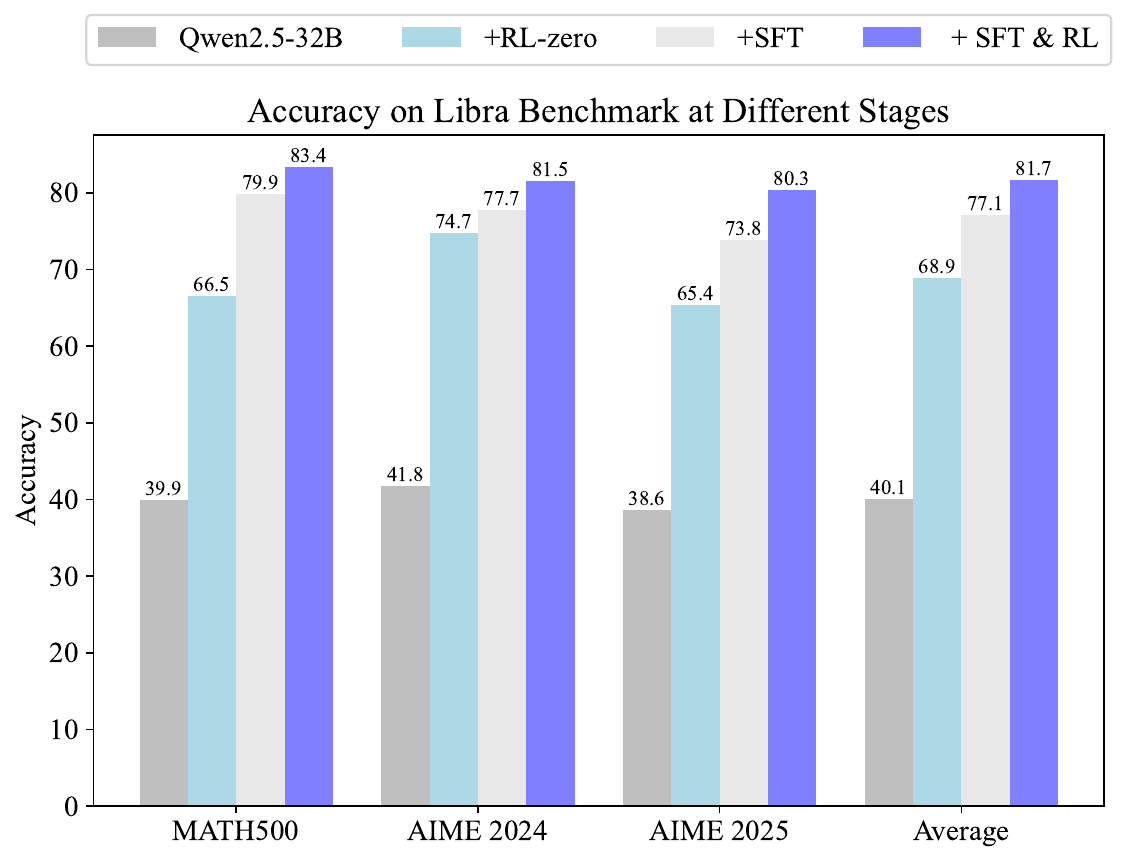}
    \label{fig:right}
  \end{subfigure}
  \caption{(a) Average accuracy on \bench during RL-zero and RL after SFT. (b) Accuracy on \bench at different stages, including initial model, RL-zero model, SFT model and SFT + RL model.}
  \label{fig:rl_curve}
\end{figure}

\paragraph{Results and Analysis} As shown in Figure~\ref{fig:rl_curve} (a), RL significantly enhances the performance of our \model-32B-MATH whether initialized from the pretrained or SFT model.
The average accuracy on \bench increases steadily and converges after approximately 400 to 500 steps in both settings.
Notably, the RL-zero variant of \modelnospace-32B-MATH, which does not utilize any distillation data, achieves an accuracy of 68.9 on the \bench, outperforming many proprietary models in Table~\ref{tabel: main_librabench}. 
This observation demonstrates the potential of training RM from scratch in entirely new settings, without relying on existing models.

However, compared to the combined SFT+RL approach, the RL-zero version converges more slowly and ultimately achieves lower final performance.  
Figure~\ref{fig:rl_curve}(b) provides a detailed comparison of the performance of the SFT checkpoint, the RL-zero checkpoint, and the RL checkpoint (trained from the SFT checkpoint), further highlighting the indispensable roles of both the SFT and RL stages in our proposed approach.

\subsection{Ablation study on Dataset Components}

\begin{figure}[t]
  \centering
  \begin{subfigure}[b]{0.49\textwidth}
    \includegraphics[width=\textwidth]{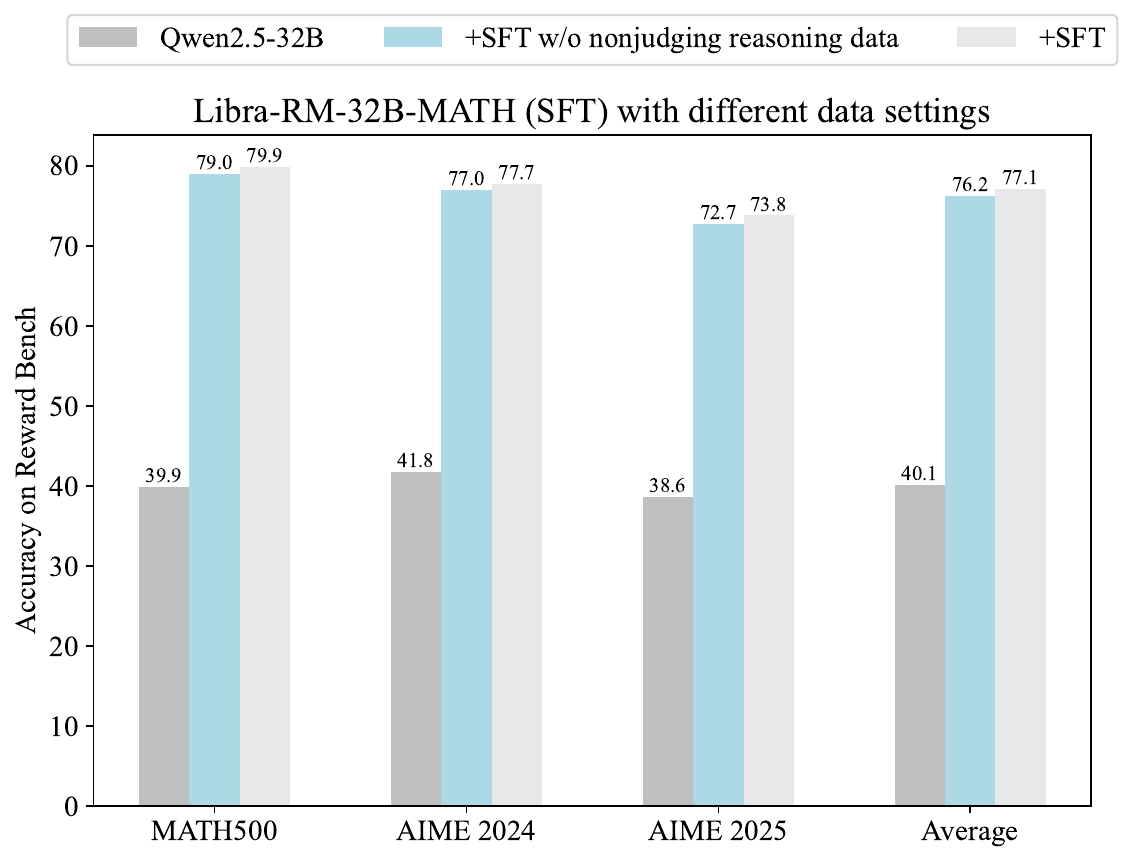}
    \label{fig:left}
  \end{subfigure}
  \hfill 
  \begin{subfigure}[b]{0.49\textwidth}
    \includegraphics[width=\textwidth]{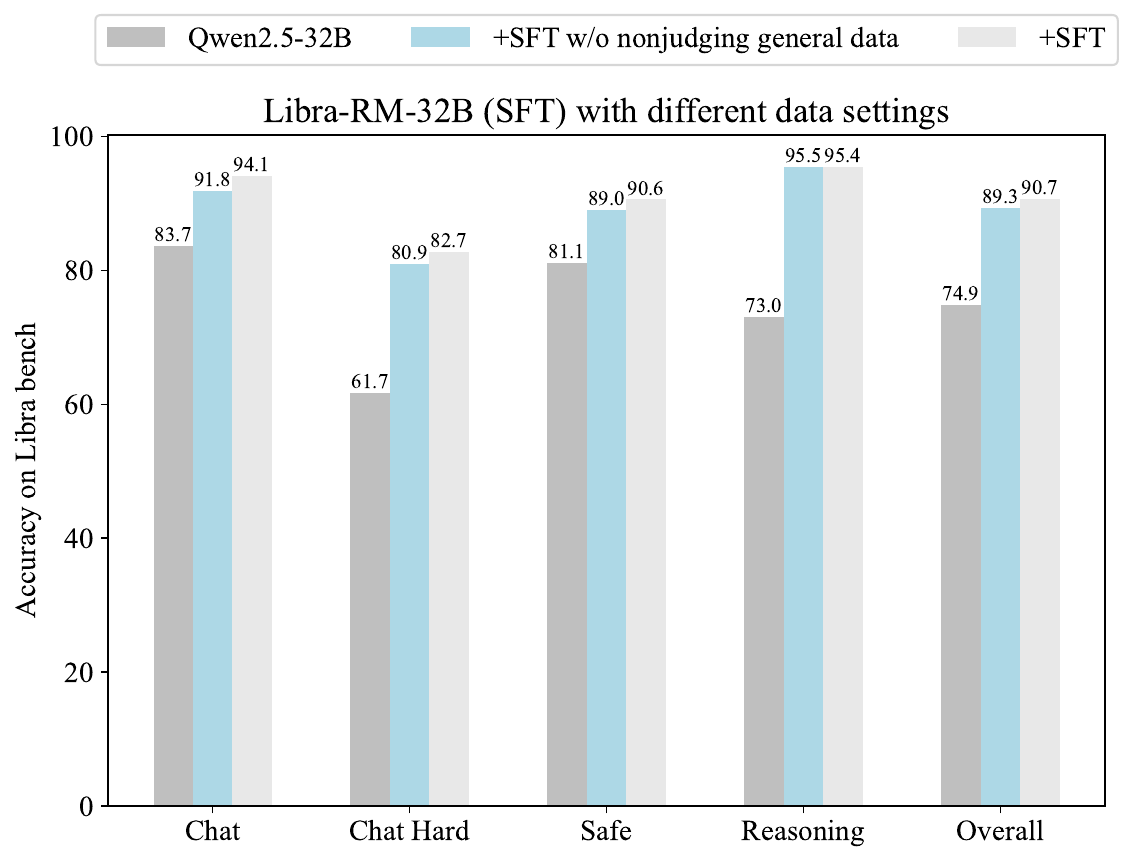}
    \label{fig:right}
  \end{subfigure}
  \caption{(a) Accuracy on \bench with different SFT data settings. (b) Accuracy on Reward Bench with different SFT data settings.}
  \label{fig:data_component}
\end{figure}

We further conduct ablation studies to assess the impact of incorporating non-judging data into the training dataset during the SFT stage, as illustrated in Figure~\ref{fig:data_component}.
\paragraph{Experimental Setups} The ablation studies are conducted based on Qwen2.5-32B, with hyperparameters set to the same values as in the \model series (detailed in Appendix~\ref{ap: train}). 
For a fair comparison, we upsample the training data for each experimental group.

\paragraph{Results and Analysis} As shown in Figure~\ref{fig:data_component}, incorporating non-judging data consistently improves the RM's performance in both reasoning and general scenarios.
Specifically, adding the non-judging reasoning data $\mathcal{D}^{rs}_{reason}$ increases the accuracy on \bench from 76.2 to 77.1, while incorporating the non-judging general data $\mathcal{D}^{rs}_{general}$ improves the accuracy on Reward Bench from 89.3 to 90.7. 
These experimental results reveal an intrinsic connection between judging and answering. The accuracy of generative reward models can be improved not only through specially designed training paradigms, but also by enhancing the model's fundamental answering abilities.

\section{Conclusions}\label{sec:con}
In this paper, we present a comprehensive framework for evaluating
and improving the performance of generative reward models in complex reasoning scenarios, introducing our \bench and \model series.
Distinct from existing RM benchmarks, the \bench is curated from a diverse collection of challenging mathematical problems and advanced reasoning models, and aims to assess pointwise judging accuracy with respect to correctness.
The \model series, including \model-32B and \model-32B-MATH, are trained through a combination of SFT and RL, where the judging process is formulated as a verifiable task.
Systematic evaluations demonstrate that our \model series achieve state-of-the-art results on various benchmarks, especially in reasoning tasks. We also provide detailed ablation studies to further validate our approach. 
Furthermore, comprehensive downstream DPO experimental results reveal the correlation between our Libra Bench and downstream application, as well as the potential of Libra-RM to further improve reasoning models with unlabeled data.

\clearpage

\bibliography{iclr2025_conference}
\bibliographystyle{iclr2025_conference}

\appendix
\clearpage
\appendix

\section{Detailed scores on RM benchmarks} \label{ap:app:eval} 
In this section, we present further details regarding our evaluation results.
Table~\ref{tab: app: reason} offers a comprehensive comparison of the performance of the \model and baseline models on the reasoning subset across RewardBench, JudgeBench, and PPE correctness metrics.
Owing to the deep thinking capacity and sophisticated training methodology, our \model series achieve state-of-the-art results on nearly every reasoning subset.

\begin{table*}[ht]
\centering
\begin{adjustbox}{width=1\columnwidth,center}
\begin{tabular}{lccccc}
\toprule
Model & Reward Bench Reasoning & PPE MATH & PPE GPQA & PPE MBPP & Judgebench  \\
\midrule
\midrule
\multicolumn{5}{c}{Discriminative Reward Model} \\
InternLM2-20B-Reward & 95.8 & 70.0 & 57.0&58.0  &63.4 \\
Skywork-Reward-Gemma-2-27B  & \textbf{98.1}  & 63.0&53.0&59.0  &64.3 \\ 
ArmoRM-8B-v0.1  & 97.3 & 71.0 & 57.0&54.0 &56.9 \\ 
Nemotron-4-340B-Reward & 93.6  & 65.0 &57.0& 49.0  & \\

\midrule
\multicolumn{5}{c}{LLM-as-a-Judge \& Generative Reward Model} \\
Claude-3.5-sonnet  & 84.7 & 86.0 & 63.0 & 54.0 & 64.3 \\
Llama-3.1-70B-Instruct & 86.0 & 73.0&56.0& 58.0 &52.3\\
DeepSeek-GRM-27B & 83.8 & 68.8 & 55.6 & 50.1 & \\
\modelnospace-32B-MATH (Ours) & 95.1  & \underline{92.8} & \underline{67.5} & \textbf{70.2} & \underline{76.6} \\
\modelnospace-32B (Ours) & \underline{97.2} & \textbf{96.3} & \textbf{71.1} & \underline{67.4} & \textbf{77.1} \\
\bottomrule
\end{tabular}
\end{adjustbox}

\caption{Detailed Scores on the reasoning subsets of existing RM benchmarks}
\label{tab: app: reason}
\end{table*}

We also present fine-grained evaluation results on Reward Bench, RMB, and PPE correctness, as summarized in Table~\ref{tab:app:RMB}, Table~\ref{tab:app:PPE}, and Table~\ref{tab:app:rewardbench}, respectively.

\begin{table*}[ht]
\centering
\small
\begin{adjustbox}{width=1\columnwidth,center}
\begin{tabular}{lccccc}
\toprule
Model & Helpful BoN & Helpful Pair & Harmful BoN & Harmful pair & Overall  \\
\midrule
\midrule
\multicolumn{5}{c}{Discriminative Reward Model} \\
InternLM2-20B-Reward & 58.5 & 76.3 & 49.9 & 67.0  & 62.9 \\
Skywork-Reward-Gemma-2-27B  & 47.2  & 65.3 & 56.1 & 72.1  &60.2 \\ 
ArmoRM-8B-v0.1  & 63.6 & 78.7 & 49.7& 66.3 & 64.6 \\ 
\midrule
\multicolumn{5}{c}{LLM-as-a-Judge \& Generative Reward Model} \\
Claude-3.5-sonnet  & 70.5 & 83.8 & 51.8 & 76.4 & 70.6 \\
Gemini-1.5-pro & 53.6 & 76.3  & 29.9 & 66.1 & 56.5  \\
Llama-3.1-70B-Instruct & 64.8 & 81.1 &55.8& 73.9 & 68.9\\
DeepSeek-GRM-27B & 62.3 & 80.5 & 57.0 & 76.1 & 69.0 \\
\modelnospace-32B-MATH (Ours) & 57.9  & 73.8 & 59.8 & 70.6 & 65.5 \\
\modelnospace-32B (Ours) & 64.8 & 79.6 & 67.5 & 79.5 & 72.9 \\
\bottomrule
\end{tabular}
\end{adjustbox}
\caption{Detailed Scores on RMB}
\label{tab:app:RMB}
\end{table*}

\begin{table*}[ht]
\centering
\small
\begin{adjustbox}{width=1\columnwidth,center}
\begin{tabular}{lcccccc}
\toprule
Model & MMLU-Pro & MATH & GPQA & MBPP-Plus & IFEval  & Mean\\
\midrule
\midrule
\multicolumn{6}{c}{Discriminative Reward Model} \\ 
InternLM2-20B-Reward & 68.0 & 70.0 & 57.0 & 58.0 & 62.0 & 63.0 \\
Skywork-Reward-Gemma-2-27B  & 54.0  & 63.0 & 53.0 & 59.0  & 54.0 & 56.6 \\ 
ArmoRM-8B-v0.1  & 66.0 & 71.0 & 57.0 &54.0 & 58.0 & 61.2 \\ 
Nemotron-4-340B-Reward & 70.0  & 65.0 &57.0& 49.0  & 63.0 & 60.8\\

\midrule
\multicolumn{6}{c}{LLM-as-a-Judge \& Generative Reward Model} \\
Claude-3.5-sonnet  & 81.0 & 86.0 & 63.0 & 54.0 & 58.0  & 68.4\\
Llama-3.1-70B-Instruct &73.0 & 73.0 & 56.0 & 58.0 & 56.0 & 63.2 \\
DeepSeek-GRM-27B & 64.8 & 68.8 & 55.6 & 50.1 & 59.8 & 59.8 \\
\modelnospace-32B-MATH (Ours) & 82.7  & 92.8 & 67.5 & 70.2 & 62.6 & 75.2 \\
\modelnospace-32B (Ours) & 86.1 & 96.3 & 71.1 & 67.4 & 65.6 & 77.3 \\
\bottomrule
\end{tabular}
\end{adjustbox}
\caption{Detailed Scores on PPE Correctness}
\label{tab:app:PPE}
\end{table*}

\begin{table*}[ht]
\centering
\small
\begin{tabular}{lccccc}
\toprule
Model & Chat & Chat Hard & Safe & Reason & Score  \\
\midrule
\midrule
\multicolumn{5}{c}{Discriminative Reward Model} \\
InternLM2-20B-Reward & 98.9 &  76.5& 89.5 & 95.8  & 90.2 \\
Skywork-Reward-Gemma-2-27B  & 96.1  & 89.9 & 93.0 &  98.1 & 94.3 \\ 
ArmoRM-8B-v0.1  & 96.9 & 76.8  & 90.5 & 97.3 & 90.4 \\ 
Nemotron-4-340B-Reward & 95.8 & 87.1 & 91.5 & 93.6  & 92.0 \\

\midrule
\multicolumn{5}{c}{LLM-as-a-Judge \& Generative Reward Model} \\
GPT-4o-0816 & 96.1 & 76.1  & 88.1&86.6 & 86.7 \\
Claude-3.5-sonnet  & 96.4 & 74.0 & 81.6 & 84.7 & 84.2  \\
Gemini-1.5-pro & 94.1 & 77.0  & 85.8& 90.2& 86.8 \\
Llama-3.1-70B-Instruct &97.2 & 70.2 & 82.8 & 86.0 & 84.0 \\
DeepSeek-GRM-27B & 94.1 & 78.3 & 88.0 & 83.8 & 86.0 \\
\modelnospace-32B-MATH (Ours) & 90.3 & 82.7 & 88.2 & 95.1 & 89.1 \\
\modelnospace-32B (Ours) & 94.7 & 86.4 & 93.3 & 97.2 & 92.9 \\
\bottomrule
\end{tabular}
\caption{Detailed Scores on Reward Bench}
\label{tab:app:rewardbench}
\end{table*}
\section{Prompt Templates} \label{sec: ap: prompts}
In subsection~\ref{sec: ap: prompts: show}, we provide the specific prompt templates utilized in our work. 
Subsection~\ref{sec: ap: prompts: exp} offers a preliminary experimental analysis of the impact of prompt design.
\subsection{Prompt Templates in our work} \label{sec: ap: prompts: show}
Figures~\ref{fig:prompt:point} and~\ref{fig:prompt:pairwise} illustrate our pointwise scoring and pairwise ranking prompt templates, respectively. The pointwise template is adapted from~\citet{wang2024helpsteer2} and~\citet{li2024crowdsourced}, while the pairwise template directly follows~\citet{li2024crowdsourced} without modification, as it has demonstrated strong performance on various tasks~\citep{frick2024evaluate}.
Figure~\ref{fig:prompt:modeleval} illustrates the outcome correctness verification prompt template, which is widely used in the curation of both our \bench and training dataset.

\subsection{Experimental Analysis}\label{sec: ap: prompts: exp}
We conduct a series of ablation studies to investigate the impact of prompt template selection. 
Table~\ref{tab:app:prompt} provides a comparison between our pointwise scoring prompt template and ``rating single response'' prompt template proposed in DeepSeek GRM~\citep{liu2025inference}.
As shown in Table~\ref{tab:app:prompt}, our pointwise scoring prompt template consistently achieves superior performance for most models, including LLM-as-a-Judge methods and specialized generative reward models.

We further perform an in-depth analysis on Qwen-32B to elucidate the observed performance differences. 
As shown in Table~\ref{tab:app:qwen3}, our prompt template substantially improves the accuracy of Qwen3-32B on incorrect samples.
We hypothesize that the Answer-Then-Compare paradigm adopted in our prompt template alleviates the confusion or interference introduced by the provided responses.

\begin{figure*}[t]
\begin{tcolorbox}[title={Pointwise Scoring Prompt Template}, colbacktitle=gray, coltitle=white, fonttitle=\bfseries, colback=gray!10!white, boxrule=0pt]

Please act as an impartial judge and evaluate the correctness of the response provided by an AI assistant to the user prompt displayed below. You will be given a user prompt and a response from AI assistant. Your job is to evaluate the quality of the response from AI assistant in terms of Correctness.\newline
Here is a detailed answer rating guide for Correctness\newline
• 4 – The response is completely correct and accurate to what is requested by the prompt with no necessary details missing and without false, misleading, or hallucinated information. The format and style of the response fully comply with the user's prompt requirements. If the prompt asks the assistant to do a task, the task is completely done and addressed in the response. \newline
• 3 – The response is mostly accurate and correct with a small amount of missing information. It contains no misleading information or hallucinations. The format and style of the response comply with the user's prompt requirements. If the prompt asks the assistant to perform a task, the task is mostly successfully attempted.\newline
• 2 – The response contains a mix of correct and incorrect information. The response may miss some details, contain misleading information, or minor hallucinations, but is more or less aligned with what the prompt asks for. The format or style of the response does not fully meet the requirements of the user's prompt. If the prompt asks the assistant to perform a task, the task is attempted with moderate success but still has clear room for improvement.\newline
• 1 – The response has some correct elements but is mostly wrong or incomplete. The response may contain multiple instances of hallucinations, false information, misleading information, or irrelevant information. The format or style of the response does not meet the requirements of the user's prompt. If the prompt asks the assistant to do a task, the task was attempted with a small amount of success. \newline
• 0 – The response is completely incorrect. All information provided is wrong, false or hallucinated. The response completely disregards the user's prompt requirements for format and style. If the prompt asks the assistant to do a task, the task is not at all attempted, or the wrong task was attempted in the response. The response is completely irrelevant to the prompt. \newline
• Check these items before evaluating Correctness: (a) Contains incorrect information (b) Contains irrelevant information (c) Key information is missing (d) Instruction is based on a false premise \newline
Begin your evaluation by generating your own answer to the prompt. You must provide your answer before judging any response.\newline
Then, compare the response from the AI assistant with your answer and evaluate the correctness of the response from AI assistant based on the answer rating guide above strictly. You must identify and correct any mistakes or inaccurate information.\newline
After providing your explanation, you must output only one of the following choices as your final verdict with a label:\newline\newline
1. The correctness score is 4: [[score=4]]\newline
2. The correctness score is 3: [[score=3]]\newline
3. The correctness score is 2: [[score=2]]\newline
4. The correctness score is 1: [[score=1]]\newline
5. The correctness score is 0: [[score=0]]\newline\newline
Example output: "My final verdict is that the correctness score is 2: [[score=2]]". \newline\newline
\texttt{<|User Prompt|>}\newline
\texttt{question}\newline\newline
\texttt{<|The Start of the AI Assistant's Response|>}\newline\verb|{answer}|\newline\texttt{<|The End of the AI Assistant's Response|>}\newline

\end{tcolorbox}
\caption{Prompt template used for pointwise scoring tasks}
\label{fig:prompt:point}
\end{figure*}

\begin{figure*}[t]
\begin{tcolorbox}[title={Pairwise Ranking Prompt Template}, colbacktitle=gray, coltitle=white, fonttitle=\bfseries, colback=gray!10!white, boxrule=0pt]

Please act as an impartial judge and evaluate the quality of the responses provided by two AI assistants to the user prompt displayed below. You will be given assistant A’s answer and assistant B’s answer. Your job is to evaluate which assistant’s answer is better.\newline\newline
Begin your evaluation by generating your own answer to the prompt. You must provide your answers before judging any answers.\newline\newline
When evaluating the assistants’ answers, compare both assistants’ answers with your answer. You must
identify and correct any mistakes or inaccurate information.\newline\newline
Then consider if the assistant’s answers are helpful, relevant, and concise. Helpful means the answer
correctly responds to the prompt or follows the instructions. Note when user prompt has any ambiguity or
more than one interpretation, it is more helpful and appropriate to ask for clarifications or more information
from the user than providing an answer based on assumptions. Relevant means all parts of the response
closely connect or are appropriate to what is being asked. Concise means the response is clear and not
verbose or excessive.
Then consider the creativity and novelty of the assistant’s answers when needed. Finally, identify any
missing important information in the assistants’ answers that would be beneficial to include when responding
to the user prompt.\newline\newline

After providing your explanation, you must output only one of the following choices as your final verdict
with a label:\newline
1. Assistant A is significantly better: [[A\verb|>|\verb|>|B]]\newline
2. Assistant A is slightly better: [[A\verb|>|B]]\newline
3. Tie, relatively the same: [[A=B]]\newline
4. Assistant B is slightly better: [[B\verb|>|A]]\newline
5. Assistant B is significantly better: [[B\verb|>|\verb|>|A]\newline\newline
Example output: "My final verdict is tie: [[A=B]]".\newline

\texttt{<||User Prompt|>}\newline
\verb|{question}| \newline\newline
\texttt{<|The Start of Assistant A's Answer|>}\newline
\verb|{answer_1}|\newline
\texttt{<|The End of Assistant A's Answer|>}\newline\newline
\texttt{<|The Start of Assistant B's Answer|>}\newline 
\verb|{answer_2}|
\newline\texttt{<|The End of Assistant B's Answer|>}

\end{tcolorbox}
\caption{Prompt template used for pairwise ranking tasks}
\label{fig:prompt:pairwise}
\end{figure*}

\begin{figure*}[t]
\begin{tcolorbox}[title={Outcome Verification Prompt Template}, colbacktitle=gray, coltitle=white, fonttitle=\bfseries, colback=gray!10!white, boxrule=0pt]

Please act as an impartial judge and evaluate the correctness of the response\'s result provided by an AI assistant to the user prompt displayed below. You will be given a user prompt, a response from an AI assistant, and a reference answer which is correct. The reference answer should be treated as authoritative, and your job is to evaluate the correctness of the response from the AI assistant by referring to the reference answer, while allowing for some flexibility in the format and structure of the response. Focus exclusively on assessing the outcome\'s validity, disregarding the response\'s derivation process. Your evaluation should follow the workflow below.\newline\newline
**Workflow**\newline
1. Extraction: Extract and output the final answer of the AI assistant to the user\'s question directly.\newline
2. Comparison: Compare the extracted final answer of the AI assistant with the reference answer. Since the reference answer is correct, the correctness of the AI assistant\'s answer depends on the consistency between the extracted final answer of the AI assistant and the reference answer. (Allowing for some flexibility in the format and structure of the answer)\newline
3. Check: Check these items for your comparison result:(a) Whether the extracted final answer has the same meaning as the reference answer. If the final answer of AI assistant\'s response is essentially equivalent to the reference answer (allowing for some flexibility in the format and structure of the response), the AI assistant\'s response should be verified as **[[Correct]]**. Otherwise, the AI assistant\'s response should be verified as **[[Wrong]]**. (b) Ensure your comparison and evaluation focus exclusively on assessing the outcome\'s validity, disregarding the response\'s derivation process.\newline
4. Output: After completing your analysis, strictly adhere to this output format:\newline

[The Final Answer of the Response]\newline
[Insert the extracted final answer here]\newline
[My Verdict]\newline
[[Correct]] or [[Wrong]]\newline\newline
**Example Output**\newline[The Final Answer of the Response]\newline x = 12 \newline \newline
[My Verdict]\newline [[Correct]]. \newline \newline
\texttt{<|User Prompt|>} \newline \verb|{question}| \newline \newline
\texttt{<|The Start of the AI Assistant\'s Response|>}\newline \verb|{answer}|\newline
\texttt{<|The End of the AI Assistant\'s Response|>}\newline\newline\texttt{<|The Start of the Reference Answer>}\newline 
\verb|{reference}|\newline
\texttt{<|The End of the Reference Answer|>}

\end{tcolorbox}
\caption{Prompt template used for correctness verification. We utilize advanced reasoning models to annotate correctness by taking the problem, response, and reference as input.}
\label{fig:prompt:modeleval}
\end{figure*}

\begin{table*}[ht]
\centering
\small
\begin{tabular}{lcc}
\toprule
Model & DeepSeek-GRM template & Our template \\
\midrule
\midrule

\multicolumn{2}{c}{LLM-as-a-Judge} \\
GPT-4o-0816& 64.2 &  65.8   \\
GPT-4.1& 69.1 &  68.5   \\
Claude-3.5-sonnet  & 59.2 &  64.7  \\
Claude-3.7-sonnet & 45.9  & 67.1  \\
Llama-3.1-70B-Instruct & 50.9 & 27.9\tablefootnote{Llama-3.1 and Claude-3.7-sonnet failed to follow the instruction requirements on some prompts, resulting in lower scores.} \\

\midrule
\multicolumn{2}{c}{LLM-as-a-Judge with thinking} \\
DeepSeek-R1 & 75.6 & 78.8 \\
Qwen3-32B & 71.8 & 78.1  \\
QwQ-32B & 73.6 & 77.5\\
R1-Distill-Qwen-32B & 59.8 & 73.7 \\
\midrule
\multicolumn{2}{c}{Generative reward models} \\
\modelnospace-32B-MATH (Ours) & 77.3 & 81.7\\
\modelnospace-32B (Ours) &  77.4 & 80.0\\
\bottomrule
\end{tabular}
\caption{\bench accuracies with different prompt templates. 
``Our template'' denotes the pointwise scoring prompt template proposed in this work, while ``DeepSeek-GRM template'' refers to the ``rating single response'' prompt template.
When applying our prompt template, judgments of our \model series are converted to the binary correctness label with a threshold of 2, consistent with the training process. For other baseline models, we select the threshold that maximizes the average accuracy to ensure a fair comparison.}
\label{tab:app:prompt}
\end{table*}

\begin{table*}[ht]
\centering
\small
\begin{tabular}{lcccc}
\toprule
Prompt Templates & MATH-500 & AIME 2024 & AIME 2025   & Average\\
\midrule
\midrule
\multicolumn{1}{c}{} & \multicolumn{3}{c}{Correct Samples} \\
Our Template & 92.7 & 88.2  & 83.7 & 88.2 \\
DeepSeek-GRM Template & 94.5 & 86.8 & 81.2 & 87.5  \\
\midrule
\midrule
\multicolumn{1}{c}{} & \multicolumn{3}{c}{Incorrect Samples} \\
Our Template & 67.8 & 68.5  & 67.8 & 68.0 \\
DeepSeek-GRM Template & 52.4 & 58.5 & 57.5 & 56.1  \\
\midrule
\midrule
\multicolumn{1}{c}{} & \multicolumn{3}{c}{All Samples} \\
Our Template & 80.2 & 78.3  & 75.8 & 78.1 \\
DeepSeek-GRM Template & 73.4 & 72.7 & 69.3 & 71.8  \\
\bottomrule
\end{tabular}
\caption{In-Depth evaluation results of Qwen3-32B on \bench using different prompt templates. 
``Our template'' denotes the pointwise scoring prompt template proposed in this work, while ``DeepSeek-GRM template'' refers to the ``rating single response'' prompt template.}
\label{tab:app:qwen3}
\end{table*}

\clearpage
\section{Details of \bench} \label{ap: bench}
We first elaborate on the annotation process in ~\ref{ap: bench: veri}, and then present several examples of our \bench in ~\ref{ap: bench: example}.

\subsection{Details of Annotation} \label{ap: bench: veri}
In summary, we utilize three approaches to annotate the outcome correctness for \benchnospace: rule-based answer matching, model-based evaluation, and human annotation. 
For model-based evaluation, we leverage reasoning models as annotators and provide them with the question, response, and reference answer to assess the correctness of the response, similar to ~\citet{seed2025seed}. The prompt template used in model-based evaluation is presented as Figure~\ref{fig:prompt:modeleval}.

For AIME 2024 and AIME 2025 where the reference answers are integers, we directly employ rule-based answer matching since it inherently achieves extremely high accuracy. We conduct a comparison between rule-based answer matching and model-based evaluation using Qwen3-32B, and observe a disagreement rate of 0.087\% between the two methods. In all instances of disagreement, the rule-based approach provides the correct annotation.

For MATH-500 level 5 where the reference answers take various forms such as integers, fractions, or expressions, we employ both model-based evaluation and human annotation to improve accuracy. 
As shown in Figure~\ref{fig:robench:me_v_rule}, conventional rule-based matching exhibits significant limitations when processing complex expressions. 
Therefore, we first adopt model-based evaluation, utilizing DeepSeek-R1, Qwen3-32B, and QwQ-32B as annotators. We observe an average disagreement rate of 0.148\%, and all samples with annotation disagreements are manually reviewed. 

To further estimate labeling accuracy on the MATH-500 subset of \benchnospace, we perform rule-based answer matching on questions whose reference answers are integers or floats. We observe only a single case of discrepancy, which is confirmed to be correctly labeled upon manual review.

\subsection{Examples} \label{ap: bench: example}
We present some examples of our \bench in this subsection. As illustrated in Figure~\ref{fig:bench:example1} and~\ref{fig:bench:example2}, each sample of \bench comprises a problem, a response, and a correctness label for the response.

\section{Experimental Details} \label{ap: train}
\subsection{Hyperparameters} 
\subsubsection{Training}
For SFT, we set the global batch size to 256 and train for 3 epochs. We utilize the AdamW optimizer with the learning rate decayed from 1e-5 to 1e-6. The warmup fraction is set to 0.03 and the clip\_gradient is set to 1.

For RL (GRPO), we set the coefficient of KL loss to 1e-3. The global batch size is also set to 256. In each rollout step, we sample 256 prompts and generate 8 responses for each prompt using a temperature of 1.0. The maximum sequence length is set to 32,768. 
We utilize the AdamW optimizer with a constant learning rate 1e-6. The warmup fraction is set to 0.03 and the clip\_gradient is set to 1. 
We adopt the Clip-Higher strategy, setting $\epsilon_{low}$ to 0.2 and $\epsilon_{high}$ to 0.28, following~\citet{seed2025seed}.
For length penalty, we set $L_{exp}$ to 16,384 and $L_{max}$ to 32,768. 
\subsubsection{Evaluation} \label{ap: eval_detail}
For deep thinking models, we sample generations with sampling parameters set to temperature=0.6 and maximum\_length=32,768. For non-thinking models,  we sample generations with sampling parameters set to temperature=0.0 and maximum\_length=4,096.

\subsection{Training Data}
For \modelnospace-32B-MATH, we combine 38,917 pointwise scoring samples in reasoning and 186,731 non-judging reasoning samples for SFT. The pointwise scoring data is curated by the V2V strategy and non-judging reasoning data is collected from in-house data and open-sourced data, including OpenR1-Math-220k and Light-R1-SFTData. In RL stage, we mix 16,874 pointwise rating samples curated by the V2V strategy and 14,591 non-judging reasoning samples from DAPO dataset~\citep{seed2025seed}.

For \modelnospace, we further expand the training data by incorporating pairwise ranking data and non-judging general data based on \modelnospace-32B-MATH. In SFT stage, we supplement 25,706 pairwise ranking samples and 26,232 in-house non-judging general samples. The pairwise ranking data is sourced from our in-house human annotations and open-sourced preference data Helpsteer2~\citep{wang2024helpsteer2}. In RL stage, we supplement 22,060 pairwise ranking samples from Skywork-Reward-Preference-80K-v0.2.

\begin{figure*}[t]
\begin{tcolorbox}[title={Example of our \benchnospace}, colbacktitle=gray, coltitle=white, fonttitle=\bfseries, colback=gray!10!white, boxrule=0pt]
\textbf{Question:}\newline
A list of positive integers has the following properties:\newline$\bullet$ The sum of the items in the list is $30$.\newline $\bullet$ The unique mode of the list is $9$.\newline $\bullet$ The median of the list is a positive integer that does not appear in the list itself.\newline Find the sum of the squares of all the items in the list.\newline\newline
\textbf{Answer:}\newline
Given the problem, we need to find a list of positive integers that satisfies the following conditions:\newline1. The sum of the items in the list is 30.\newline2. The unique mode of the list is 9.\newline3. The median of the list is a positive integer that does not appear in the list itself.\newline\newline After extensive trial and error, we considered various configurations of the list. The key steps and considerations were:\newline\newline1. **Mode Requirement**: The number 9 must appear more frequently than any other number.\newline2. **Median Requirement**: The median must be an integer not present in the list, implying the list must have an even number of elements.\newline3. **Sum Requirement**: The total sum of the list must be 30.\newline\newline After multiple attempts, the closest valid configuration found was the list $[1, 2, 4, 5, 9, 9]$, though its median is 4.5, which is not an integer. However, due to the constraints and the requirement for the mode to be 9, this configuration was considered the closest valid solution despite the median issue. \newline\newline The sum of the squares of the items in this list is calculated as follows:\newline$1^2 + 2^2 + 4^2 + 5^2 + 9^2 + 9^2 = 1 + 4 + 16 + 25 + 81 + 81 = 208$\newline\newline Thus, the final answer is:\newline$\boxed{208}$\newline\newline
\textbf{Reference:}\newline236\newline\newline
\textbf{Label:}\newline0 (Incorrect)\newline
\end{tcolorbox}
\caption{An example from our \bench}
\label{fig:bench:example1}
\end{figure*}

\begin{figure*}[t]
\begin{tcolorbox}[title={Example of our \benchnospace}, colbacktitle=gray, coltitle=white, fonttitle=\bfseries, colback=gray!10!white, boxrule=0pt]

\textbf{Question:}\newline
Let $A$ be the set of positive integer divisors of $2025$. Let $B$ be a randomly selected subset of $A$. The probability that $B$ is a nonempty set with the property that the least common multiple of its elements is $2025$ is $\frac{m}{n}$, where $m$ and $n$ are relatively prime positive integers. Find $m + n$.\newline
\textbf{Answer:}\newline
To solve the problem, we start by noting that 2025 factors into primes as $3^4 \times 5^2$. The set $A$ of its positive divisors has $(4+1)(2+1) = 15$ elements. We need to find the probability that a randomly selected non-empty subset $B$ of $A$ has an lcm of exactly 2025.\newline\newline First, we use inclusion-exclusion principle to count the number of subsets whose lcm is exactly 2025. For the lcm to be 2025, the subset must include at least one divisor with $3^4$ and at least one divisor with $5^2$.\newline\newline1. **Total subsets of $A$**: $2^{15}$.\newline2. **Subsets lacking $3^4$**: These subsets can only include divisors with exponents of 3 up to 3. There are $4 \times 3 = 12$ such divisors, so there are $2^{12}$ subsets.\newline 3. **Subsets lacking $5^2$**: These subsets can only include divisors with exponents of 5 up to 1. There are $5 \times 2 = 10$ such divisors, so there are $2^{10}$ subsets.\newline4. **Subsets lacking both $3^4$ and $5^2$**: These subsets can only include divisors with exponents of 3 up to 3 and exponents of 5 up to 1. There are $4 \times 2 = 8$ such divisors, so there are $2^8$ subsets.\newline\newline Using inclusion-exclusion, the number of subsets with lcm exactly 2025 is:\newline$2^{15} - 2^{12} - 2^{10} + 2^8$\newline Calculating each term:\newline$2^{15} = 32768, \quad 2^{12} = 4096, \quad 2^{10} = 1024, \quad 2^8 = 256$\newline $32768 - 4096 - 1024 + 256 = 27904$\newline\newline The total number of subsets of $A$ is $2^{15} = 32768$. The probability is thus:\newline$\frac{27904}{32768} = \frac{109}{128}$\newline where the fraction is simplified by dividing both numerator and denominator by their greatest common divisor, $2^8$.\newline\newline Thus, $m = 109$ and $n = 128$, leading to $m + n = 109 + 128 = 237$.\newline\newline$\boxed{237}$\newline\newline
\textbf{Reference:}\newline237\newline\newline
\textbf{Label:}\newline1  (Correct)\newline
\end{tcolorbox}
\caption{An example from our \bench}
\label{fig:bench:example2}
\end{figure*}

\begin{figure*}[t]
\begin{tcolorbox}[title={Comparison between rule-based answer matching and model-based evaluation}, colbacktitle=gray, coltitle=white, fonttitle=\bfseries, colback=gray!10!white, boxrule=0pt]
\textbf{\large{Case 1:}}\newline\newline
\textbf{Answer:}\newline
......\newline
Checking that these solutions do not make any denominators zero in the original equation confirms they are valid. Therefore, the real roots are:\newline\newline $\boxed{1 + \sqrt{19}}$ and  $\boxed{1 - \sqrt{19}}$\newline\newline
\textbf{Reference:}\newline$1\pm\sqrt{19}$\newline\newline
\textbf{Rule-Based Label:}\newline0\newline\newline
\textbf{Model-Based Label:}\newline1\newline\newline

\textbf{\large{Case 2:}}\newline\newline
\textbf{Answer:}\newline
......\newline
Thus, the set of all possible values of the expression $\frac{c}{a} + \frac{a}{b + c} + \frac{b}{c}$ is: \newline\newline$\boxed{[2, \infty)}$ \newline\newline
\textbf{Reference:}\newline$(2,\infty)$\newline\newline
\textbf{Rule-Based Label:}\newline1\newline\newline
\textbf{Model-Based Label:}\newline0\newline\newline
\textbf{\large{Case 3:}}\newline\newline
\textbf{Answer:}\newline
......\newline
Converting this to degrees:\newline         $\frac{3\pi}{14} \times \frac{180^\circ}{\pi} = \frac{540^\circ}{14} = \frac{270^\circ}{7}$\newline\newline Thus, the degree measure of angle $\angle BAC $ is $\boxed{ \texttt{\\dfrac\{270\}\{7\}}}$. \newline\newline
\textbf{Reference:}\newline$\verb|\frac{270}7|$\newline\newline
\textbf{Rule-Based Label:}\newline0\newline\newline
\textbf{Model-Based Label:}\newline1\newline
\end{tcolorbox}
\caption{Comparison between rule-based answer matching and model-based evaluation}
\label{fig:robench:me_v_rule}
\end{figure*}

\end{document}